# A Digital Twin Framework Utilizing Machine Learning for Robust Predictive Maintenance: Enhancing Tire Health Monitoring

**Vispi Karkaria**
Department of Mechanical Engineering,
Northwestern University,
2145 Sheridan Road,
Evanston, IL 60208
e-mail: vispikarkaria2026@u.northwestern.edu

**Jie Chen**
Department of Mechanical Engineering,
Northwestern University,
2145 Sheridan Road,
Evanston, IL 60208
e-mail: jie.chen@northwestern.edu

**Christopher Luey**
Department of Mechanical Engineering,
Northwestern University,
2145 Sheridan Road,
Evanston, IL 60208
e-mail: christopher.luey@u.northwestern.edu

**Chase Siuta**
Michelin North America,
Greenville, SC 29615
e-mail: chase.siuta@michelin.com

**Damien Lim**
Michelin North America,
Greenville, SC 29615
e-mail: damien.lim@michelin.com

**Robert Radulescu**
Michelin North America,
Greenville, SC 29615
e-mail: robert.radulescu@michelin.com

**Wei Chen[1]**

Department of Mechanical Engineering,

Northwestern University,

2145 Sheridan Road,

Evanston, IL 60208

e-mail: weichen@northwestern.edu

## ABSTRACT

*We introduce a novel digital twin framework for predictive maintenance of long-term physical systems. Using monitoring tire health as an application, we show how the digital twin framework can be used to enhance automotive safety and efficiency, and how the technical challenges can be overcome using a three-step approach. Firstly, for managing the data complexity over a long operation span, we employ data reduction techniques to concisely represent physical tires using historical performance and usage data. Relying on these data, for fast real-time prediction, we train a transformer-based model offline on our concise dataset to predict future tire health over time, represented as Remaining Casing Potential (RCP). Based on our architecture, our model quantifies both epistemic and aleatoric uncertainty, providing reliable confidence intervals around predicted RCP. Secondly, to incorporate real-time data, we update the predictive model in the digital twin framework, ensuring its accuracy throughout its life span with the aid of hybrid modeling and the use of discrepancy function. Thirdly, to assist decision making in predictive maintenance, we implement a Tire State Decision Algorithm, which strategically determines the optimal timing for tire replacement based on RCP forecasted by our transformer model. This approach ensures our digital twin accurately predicts system health, continually refines its digital representation, and supports predictive maintenance decisions. Our framework effectively embodies a physical system, leveraging big data and machine learning for predictive maintenance, model updates, and decision-making.*

---

[1] Corresponding author information can be added as a footnote.

## 1. INTRODUCTION

Digital twins, which are virtual counterparts of physical entities through mutual communications, enable the forecasting of failures, prevention of unforeseen damages, and optimization of maintenance routines, thereby improving both the lifespan and maintenance strategies of various systems [1]. This technology is increasingly applied across sectors such as manufacturing, aerospace, and construction due to its proficiency in offering real-time damage assessments and facilitating maintenance decision-making [2]. Specifically, in the realm of vehicle maintenance, digital twin technology signifies a paradigm shift by enabling real-time monitoring and predictive analytics concerning future tire conditions through the application of advanced machine learning techniques [3].

Digital twin is in line with the principles of a circular economy—emphasizing resource efficiency and waste minimization—the significance of extending tire life cannot be overstated  [4]. Given their impact on vehicle fuel efficiency and safety, tires notably influence both operational expenses and environmental sustainability [5] [6]. The growing concern over tire waste, projected to reach between 1.3 and 1.5 billion tons annually by 2025, underscores the urgency for solutions that mitigate efficiency losses and material degradation under diverse operating conditions [7]. By offering precise predictions of tire durability against the backdrop of changing road conditions, weather, load variances, driving habits, and maintenance routines, digital twins play a pivotal role [8] [9] [10]. They not only improve maintenance scheduling to ensure safety but also diminish environmental impact by reducing the need for frequent tire replacements [11]. Consequently, these predictive capabilities facilitate the optimized utilization of tires and endorse sustainable practices, aligning with the core objectives of a circular economy [12].

In the realm of digital twins, the predictive maintenance sector confronts numerous obstacles, including data acquisition, model training, uncertainty quantification/propagation, model verification, model update and decision making. A hurdle is the collection of comprehensive operational, telematics and usage data, essential for building a digital twin that reflects product or system behavior in real-world scenarios [13]. The process of training surrogate models in the digital twin to estimate predictive maintenance schedules involves processing complex datasets to understand patterns indicative of wear and deterioration. This task demands advanced algorithms and substantial computing power, coupled with the capability to transmit and continuously collect data in real-time via the Internet of Things (IoT) [13] [14].

In predictive maintenance, machine learning techniques significantly enhance digital twin functionality. Optimal trees provide accurate maintenance predictions by dynamically updating digital twins with relevant physics-based models [15]. Kernel-based regressions train surrogate models, reducing computational demands while maintaining accuracy [16]. Reinforcement learning improves adaptability and precision, enabling generalization across various setups [17]. Scientific machine learning combines model order reduction with dynamic, intelligent computing [18]. Lastly, probabilistic machine learning crucially manages uncertainties in updates, ensuring long-term reliability [19]. However, these methods, including optimal trees and kernel-based regressions, may not suit applications requiring deep learning-based time series analysis, as they struggle to model complex temporal dependencies crucial for forecasting in large-scale data.

Understanding and incorporating uncertainty into the digital twin is critical. It involves quantifying the inherent variability in usage data, thus enabling the construction of digital twins that offer more precise and dependable forecasts of tire health [20]. This understanding of potential outcomes and the confidence in the predictions supports more informed maintenance decisions, enhancing tire longevity [21]. Furthermore, validating these digital twins to ensure their prediction correspond with real tire health necessitates thorough evaluation and adjustment, highlighting the complexity of leveraging digital twins for tire health assessment.

Addressing challenges related to model updates and decision-making is fundamental in refining predictive maintenance [22]. These tasks require seamless integration of real-time data, adaptive model calibration, and informed decision-making based on maintenance predictions. For instance, using digital twins in maintenance frameworks can mitigate operational risks and cut costs, underscoring the value of timely and historical data in enabling dynamic decision-making [23]. Yet, the challenge of sparse data over a long term life span complicates the model update process, affecting the digital twin's decision-making reliability [24]. Decision-making becomes challenging when damage indicators are difficult to measure, due to the variability and unpredictability of usage conditions [24,25]. These concerns underline the urgent need for advanced computational methods and analytics to navigate limited data situations for model updates and enhance decision-making capabilities, even in the absence of direct damage signs [26].

In this paper, we address the challenges in digital twin technology for products with long-term operations using tire health monitoring as an example and introduce an approach to model updating and decision-making. Our methodology involves the employment of discrepancy-based model update algorithms, designed to effectively learn from limited datasets collected during the lifespan of operations. At the core of our digital twin framework is the Temporal Fusion Transformer (TFT), an attention-based time-series model that employs time series analytics, essential for understanding sequential data, to accurately interpret indicators of tire health and damage. Unlike traditional Transformer, RNN, or LSTM models, the Temporal Fusion Transformer was selected for its ability to handle multiple input features with varying temporal dependencies, allowing for a more nuanced integration of both static and time-varying inputs, which is critical in modeling the complex interactions in tire health monitoring under varied operational conditions [27]. This strategy affords a deeper insight into tire behavior across large operating conditions, enabling more precise estimations of tire health. Furthermore, the framework integrates a decision-making algorithm, termed the tire state decision algorithm, which utilizes the model's RCP prediction to make the decision for tire replacement, thus addressing the inherent variability and unpredictability of tire failure. This approach enhances the reliability and precision of digital twin systems in monitoring tire conditions and facilitating efficient tire health management.

This paper is structured as follows: Section 2 delves into the discussion of the digital twin framework for predictive maintenance, and the challenges faced in building a digital twin. Section 3 outlines the three important stages involved in constructing the health digital twin using tire as an application example. Firstly, we delve into the offline model training phase, where the TFT tailored for tire health is trained using historical data. Next, we describe the model update algorithm, which applies a discrepancy-based method to refine the model with new data, resulting in an enhanced hybrid model. Lastly, we introduce the tire state decision algorithm, which is designed to interpret the updated model's outputs to guide timely tire maintenance decisions. In Section 4, we study the verification process for the Tire Health TFT trained offline, assessing the tire health digital twin's capability in self-updating and its real-time decision-making ability regarding the optimal timing for tire replacements. Finally, we conclude with a discussion on the implications of our findings for the future of tire maintenance using digital twin technology.

## 2. FRAMEWORK OF A DIGITAL TWIN FOR PREDICTIVE MAINTENANCE

In this section, we discuss the concept of a digital twin and its role in predictive maintenance applications. We begin by stating the definition of a digital twin, and then detailing its composition and functionality within the context of predictive maintenance. Furthermore, we address the four main challenges encountered in developing a digital twin for predictive maintenance applications.

A digital twin is a set of virtual information constructs that mimics the structure, context, and behavior of a natural, engineered, or social system (or system-of-systems), is dynamically updated with data from its physical twin, has a predictive capability, and informs decisions that realize value. This virtual model is continuously updated with data throughout its lifecycle, guiding decisions that maximize the value of the physical asset(s) [20]. In the context of predictive maintenance, the digital twin serves as a dynamic tool for monitoring, analyzing, and optimizing the performance and maintenance of physical assets. The core components of a digital twin for predictive maintenance include real-time data acquisition, historical data analysis, predictive modeling, model-based update, and decision-making algorithms. This seamless integration allows for the constant update of the surrogate model, reflecting the current state of the physical counterpart accurately, as shown in Figure 1. To demonstrate a general framework of digital twin for predictive maintenance, we use the example of the Tire health digital twin in this paper.

In Figure 1, the digital twin begins with the offline training of the Tire Health Temporal Fusion Transformer (TFT) model in Step 1, leveraging historical datasets which has operating parameters ($T_t$) - conditions under which the tire operates, usage parameters ($U_t$) - how the tire is used, and state parameters ($M_t$) - the current condition of the tire. Additionally, within our digital twin framework, we get our dataset with inputs derived from a physics-based Tire Design Finite Element Method (FEM) integrating physical insights with measured data. Incorporating a physics-based Tire Design Finite Element Method (FEM) is crucial to accurately understand the tire's physics-based state, ensuring a comprehensive analysis of its condition through the integration of physical principles

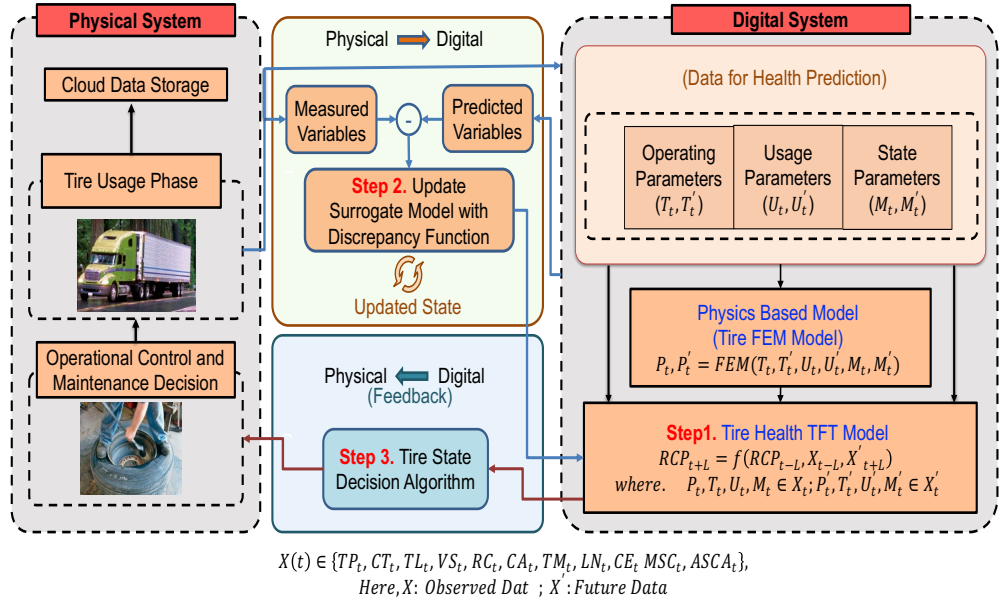

**FIGURE 1.** Tire health digital twin framework demonstrating the flow of information, and important components of digital twin like offline training, model update and decision making. The operating, usage and state parameters $(T_t, U_t, M_t)$ collectively form the input variable $(X)$ for the tire health TFT model.

with observed data. Then the Tire Health TFT model, a critical component of the Tire Health Digital Twin, facilitates real-time predictions of the damage state. A continuous quantity, named as Remaining Casing Potential (RCP), is considered as the damage state parameter). RCP serves as a key indicator of tire endurance damage, allowing for proactive maintenance decisions. The predictions by the Tire Health TFT model are subsequently compared with real-world instances of tire damages. This comparison allows us to quantify the discrepancy, effectively measuring the difference between the model's predictions and the actual tire damage data in Step 2. We utilize observed discrepancies to refine our Tire Health TFT model. It is important to highlight that, following an update, our model evolves into a hybrid version. Despite this transformation, we continue to refer to it as the Tire Health TFT model for consistency and clarity in our discussion in this paper. Then the updated Tire Health TFT model, with the Tire State

Decision Algorithm in Step 3, informs timely tire replacement decisions. Thus, our tire health digital twin has the surrogate model, which is updated in real-time, and aids in making predictive maintenance decisions.

Moving forward we address four research challenges in building a tire health digital twin framework. The first challenge involves performing real-time tire health prediction while the tire is in use. Our hypothesis suggests that through an offline-trained machine learning model based on historical tire health data integrated with real usage data within the digital twin framework, we can make online health predictions.

The second challenge deals with the imbalance in predictive maintenance data where data indicating damage is substantially less than non-damage data. To tackle this, we collect damage tire health data through the Random Forest Model which effectively predicts the RCP by learning from the imbalance in the data [28]. In our previous paper to address the challenge of data imbalance, particularly the scarcity of data representing damage, we employ a variance reduction SMOTE (Synthetic Minority Over-sampling Technique) method. This approach enhances the model's learning capacity by generating synthetic examples of the underrepresented damage data, thus balancing the dataset for more effective and unbiased RCP prediction. Although the data collected from the Random Forest model, specifically RCP, is not inherently time-series in nature, it encapsulates temporal dynamics by reflecting the evolving condition of tires over operational mileage. This allows us to reinterpret RCP values as a time-series, enabling the training of a TFT to capture complex temporal relationships and predict future tire conditions with higher accuracy, leveraging both the historical and current states of tire wear.

The third challenge we address involves updating the model in real-time. In the context of predictive maintenance, a significant hurdle is the time-intensive process of collecting update data, as endurance damages and significant wear often happens over extended periods of operation. This delay in observing critical changes challenges the timely adaptation of the model, emphasizing the need for a strategy that can effectively leverage limited yet important data to maintain digital twin. This real-time update ensures the model's outputs remain both precise and applicable, enhancing its utility in practical settings [29]. We propose a hybrid modeling approach that accounts for measurement discrepancy—known as residuals—which are crucial for fine-tuning the digital twin's accuracy. Discrepancy, the differences between observed and predicted values, are key indicators of model performance and help in adjusting the model to reflect the dynamic nature of real-world conditions [30]. Furthermore, the discrepancy approach allows for updating the model with fewer datasets by efficiently identifying critical deviations that necessitate adjustment, thereby focusing on the most impactful data points for refinement rather than requiring large volumes of new information.

The fourth research question addresses the importance of robust decision-making in the presence of uncertainty within predictive maintenance [31]. Our proposed solution involves an offline-trained machine learning model that quantifies both epistemic and aleatoric uncertainty. Epistemic uncertainty, or the uncertainty in the model due to limited data, is reduced through learning from extensive historical data, whereas aleatoric uncertainty, which stems from inherent variability in the data, is quantified to provide a measure of confidence in the predictions [32,33]. This dual approach enables more precise decision-making by providing a clearer picture of the tire's health and the likelihood of endurance damages, thus optimizing maintenance schedules, and extending tire life. Additionally, the continuous model update capability of the digital twin aids in reducing the aleatoric uncertainty in tire health monitoring because it dynamically incorporates real-time operational data and feedback into the predictive model. This process not only refines the model's predictions over time but also helps in uncovering and understanding previously unknown patterns or 'unknown unknowns' in the data [34]. By using to new collected data as they are observed, the tire health digital twin becomes progressively more accurate and robust, effectively narrowing the gap of uncertainty and enhancing the reliability of maintenance decisions under variable operational conditions. Robust decision-making is essential in predictive maintenance because it ensures that maintenance actions are both timely and cost-effective, preventing unnecessary downtime and extending the lifespan of equipment under fluctuating conditions.

Overall, the tire health digital twin serves as a tool in predictive maintenance, allowing for enhanced tire health prediction. In the next section we describe the different steps behind building the tire health digital twin.

### 3. BUILDING THE TIRE HEALTH DIGITAL TWIN

This section outlines the steps for implementing the tire health digital twin. We begin with data collection and data handling processes, establishing the foundation of our framework. Subsequently, we delve into the steps of the machine learning model designed to estimate the tire's RCP, thereby predicting its state of life. Finally, we discuss about the mechanisms for updating the tire health digital twin and the algorithm to find the optimal time to change the tire.

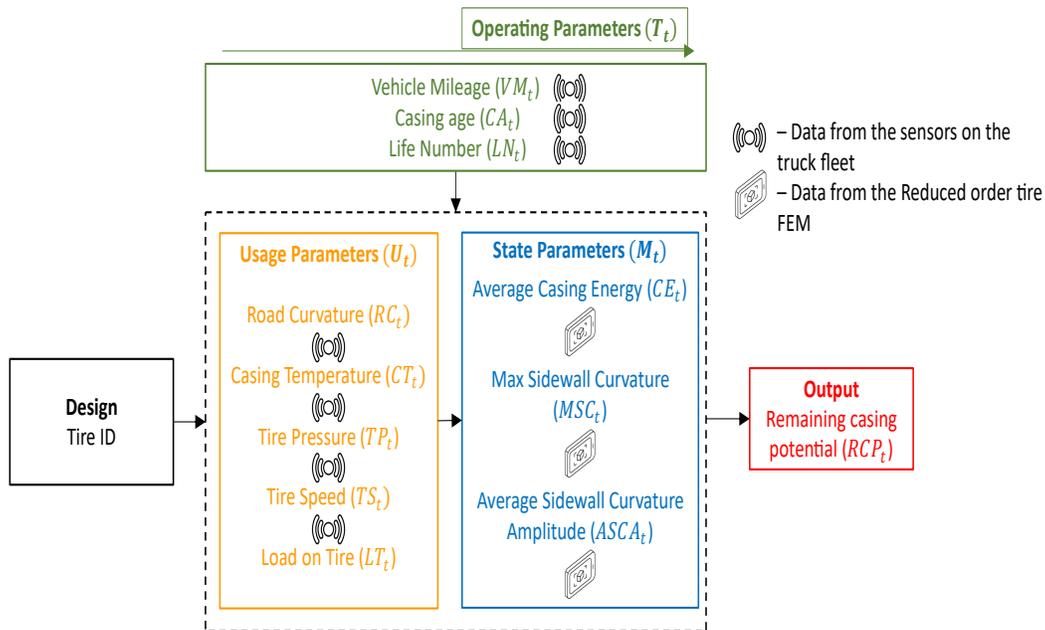

**FIGURE 2.** Flowchart depicting the causal relationships and information flow among input variables leading to the tire health TFT output. The diagram illustrates the hierarchically structured data inputs, from tire mileage to individual tire factors, culminating in the prediction of RCP.

### 3.1 Data collection and data handling for offline training in the tire health digital twin

In our study, we collected a dataset of 2506 tire life data points, which are structured as time series and have been collected through sensors within the vehicle and tires and through the Tire Finite Element Model (FEM). These historical data points captured information regarding the tires' usage conditions and state at various timestamps of tire life, including variables such as road curvature of the tire ($RC_t$), casing temperature ($CT_t$), tire pressure ($TP_t$), vehicle speed ($VS_t$), and load on tire ($LT_t$). Moreover, the dataset incorporates metrics related to the usage state of the casing, like average casing energy ($CE_t$), average sidewall curvature amplitude ($ASCA_t$) and maximum sidewall curvature ($MSC_t$) [28]. By integrating these variables with metadata such as tire mileage ($TM_t$), casing age ($CA_t$), and life number ($LN_t$), our machine learning model processes the time in service to produce a predictive output: the RCP. More details about the definition of the input variables to the Tire health TFT are given in the appendix at the end of the paper. RCP which is the output metric of the Tire health TFT is collected through our random forest model which we proposed in our previous paper [28]. This output serves as an indicator for understanding tire endurance damage and forecast the optimal timeline for maintenance or replacement, thereby enhancing the safety and efficiency of vehicle operation. The causal effect relationship between different variables is shown in Figure 2, where all the eleven variables are classified into operating, usage, and state parameters as a reference to Figure 1. The causal effect relationships illustrated in Figure 2 are critical for the construction of our predictive maintenance digital twin, as they guide the feature selection and modeling process. By understanding these relationships, we can prioritize the variables that most significantly impact tire health and durability, enabling the digital twin to focus on predictive insights that drive maintenance decisions and improve operational efficiency.

In our dataset, each tire is represented by a time series of its life containing between 25,000 to 30,000 data points. Multiplying this by the number of distinct tire time series results in a total of 76,789,042 data points. Training a machine learning model specifically our Tire Health TFT with such a voluminous dataset can significantly decelerate the training process [35]. Moreover, it is often unnecessary to use all data points for effective model training, as this can lead to a phenomenon known as 'overfitting,' where the model learns the noise in the data rather than the underlying pattern we wish to model [35]. Reducing the dataset to the most informative features and instances can lead to a more generalized model that performs better on unseen data. To address this, we employ a data reduction technique that ensures the essential information and the covariance among the time series covariates are preserved while reducing the dataset size. This approach optimizes the training efficiency of the machine learning model without compromising the quality of the predictive model.

The foundation of our technique is the application of Gaussian kernel smoothing, a well-established method for suppressing noise and highlighting trends in time series data [36]. The tire health data collected from the field often contains a significant amount of noise or uncertainty, necessitating robust methods like Gaussian kernel smoothing to effectively suppress noise and enhance the visibility of underlying trends [37]. We define our kernel $k$ as a normalized Gaussian function for gaussian kernel smoothening:

$$k(z) = \frac{1}{\sqrt{2.\pi.\sigma^2}} . e^{-\frac{z^2}{2.\sigma^2}} \tag{1}$$

Where $z$ represents the points in our discretized domain, and $\sigma$ is the standard deviation, dictating the width of the kernel. In the context of tire life prediction, the variable $z$ in the Gaussian kernel function can be interpreted as the discrete time or mileage intervals at which tire condition data is recorded, while the standard deviation $\sigma$ controls the smoothing effect, determining how much influence neighboring data points have on the smoothed value at $z$ [38]. The discretized and normalized kernel $k$, when convolved with our time series data $D_i$, yields a smoothed representation $s_i$:

$$s_i = D_i * k(z) \qquad (2)$$

Post-smoothing, we proceed to the core of our data reduction strategy—adaptive down sampling. This step is crucial in distilling the dataset to its most informative components. By calculating the first-order difference $\Delta s_i$ to approximate the rate of change in the smoothed data, we can identify and retain pivotal points in the series that signify substantial variations in the data, a process mathematically depicted as:

$$\Delta s_i = |s_i[t+1] - s_i[t]| \qquad (3)$$

We aggregate these rates of change across all covariates at each time point $t$ to form a collective change metric $A_t$. Points where $A_t$ surpasses a predefined threshold $\theta$ are preserved:

$$A_t = \sum_{i=1}^{n} \Delta s_{it} \qquad (4)$$

By setting a threshold $\theta$, we can filter the dataset. Points where $A_t$ exceeds $\theta$ are indicative of significant changes in the data and are retained. Consequently, the reduced dataset $D_{\mathrm{reduced}}$ is constructed by selecting data points from the original dataset $D$ based on the condition that $A_t > \theta$:

$$D_{reduced} = \{D_i \in D | A_t(D_i) > \theta\} \tag{5}$$

By implementing this data reduction technique, we effectively streamline the dataset from over 76 million points to a manageable size of 365K points that conveys the same analytical value. Each Tire Health Time Series Data has 150-250 datapoints. This method ensures that the machine learning models are provided with high-quality, representative data, fostering both the accuracy of predictive analytics and the expediency of computational processes. In the next section, we will dig deeper into our surrogate model which is the tire health TFT for our tire health digital twin.

### 3.2 Surrogate model offline training for the tire health digital twin:

In the development of the tire health digital twin, the TFT was selected as the surrogate model for the tire health digital twin for its capacity to analyze complex time series data. The TFT is selected because of its architecture that combines time-varying inputs with static factors, a feature important for accurately capturing the multidimensional patterns of tire endurance damage and overall performance in predictive maintenance [27] [39]. This model stands out for its use of attention mechanisms and variable selection networks, which intelligently identify and focus on the most critical factors affecting tire life, such as tire internal energy, load capacity, and environmental conditions [40]. Such feature prioritization ensures the model's forecasts are both accurate and relevant by using fewer time sequences with more input variables. Additionally, by forecasting key indicators of tire health, which is RCP, the TFT provides insights for preemptive maintenance actions and operational adjustments.

As shown in Figure 3. The tire health TFT model takes in Offline observed data and offline future data as input and predicts RCP sequence as output. The TFT model's architecture is designed to channel input information through a series of computational layers, each serving a unique role in the predictive process. Initially, a variable selection network identifies the most relevant input variables in time for prediction. These selected variables are then encoded and decoded through LSTM networks to capture temporal dependencies. Subsequently, GRN refine the feature representation, which, in conjunction with a multi-head attention mechanism, allows the model to focus on critical

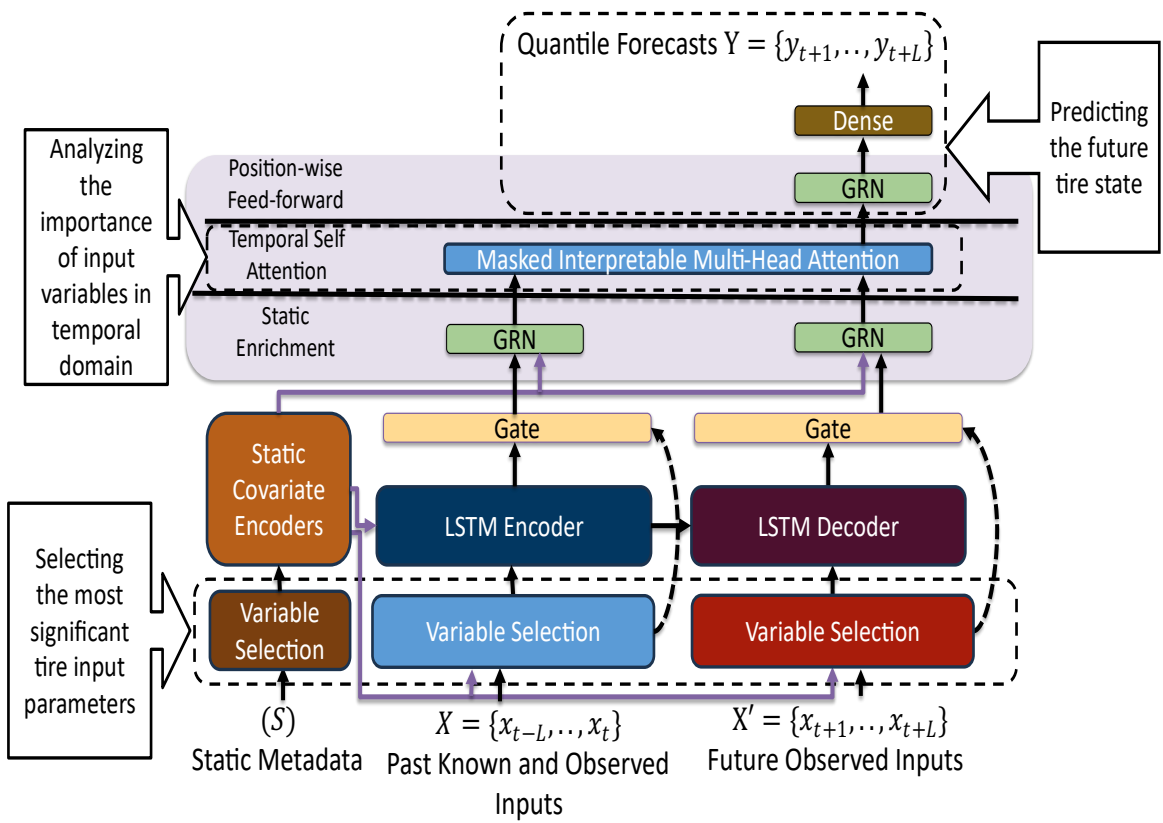

**FIGURE 3.** Architecture of the tire health TFT model. It illustrates the multilayered flow of information within the model, starting from variable selection at the lowest level, ascending through the LSTM encoder-decoder structure for temporal analysis, and culminating in the dense and GRN layers that synthesize the input to predict the RCP.

time-dependent patterns. The final stage employs a dense layer to integrate these insights and output the predicted RCP sequence, providing a robust forecast of tire health. In input and output data for tire health TFT are as follows:

The predicted ($RCP_{t+L}$), is a function of past RCP values ($RCP_{t-L}$), past observed data ($X_{t-L}$), and future data ($X'_{t+L}$):

$$RCP_{t+L} = f(RCP_{t-L}, X_{t-L}, X'_{t+L}) \qquad (6)$$

where,

$Offline\ Observed\ Data$ ($X$): The historical tire health data comprising the series $X = \{x_{t-L+1}, \dots, x_t\}$, where $t$ is denotes the current time step and $L$ is the window size for historical data.

$Offline\ Future\ Data$ ($X'$): The projected tire health data comprising of covariance $X' = \{x'_{t+1}, \dots, x'_{t+L}\}$, with $t$ as the current time step and $L$ as the prediction horizon.

$RCP\ Sequence$ ($RCP$): The forecasted RCP values are described by the sequence $RCP = \{RCP_{t+1}, \dots, RCP_{t+L}\}$, which represent the tire health TFT's output over prediction horizon.

As we study the stages of information flow in the tire health TFT model, we begin with the variable selection network. This network identifies and prioritizes the most critical input features, discerning both static and time-variant data crucial for predicting the tire's future health condition. It employs a gating mechanism, filtering the data to retain only the most significant variables for subsequent analysis. The dynamic nature of this network ensures that at each time step, the selected historical and future data are optimally weighted to reflect their importance in estimating the RCP. This layer of the tire health TFT model focuses on key indicators such as tire pressure, temperature, and tread depth, utilizing their time series data to accurately forecast RCP.

Following the variable selection stage in the tire health TFT model, we proceed to the LSTM encoder-decoder structure, which is fundamental for capturing and interpreting the tire health data. The encoder analyzes the selected variables at each timestep, embedding the temporal patterns and historical context into a compact form. This encoding serves as a chronological summary, encapsulating the progression of tire health indicators over time.

Subsequently, the LSTM decoder extrapolates these temporal patterns, using the distilled historical data to anticipate the future trajectory of tire health, specifically predicting the RCP. By doing so, the model synthesizes past trends to generate future projections, enabling it to anticipate the point at which a tire may reach critical wear levels. This sequential processing is central to understanding how a tire's condition evolves and determining the most opportune moments for maintenance interventions.

After the LSTM encoder and decoder have processed the selected variables to capture and project the temporal patterns, the Tire Health TFT model employs a GRN (Gated Residual Network) to refine the outputs. The GRN is important in the model for integrating non-linear transformations and providing a means to learn from the residuals—the differences between predicted and actual values. This is especially pertinent for tire life prediction, where it's crucial to understand not just the direct patterns, but also the nuances and exceptions in the data which can indicate more subtle forms of wear or imminent damage.

Simultaneously, the model incorporates a Masked Interpretable Multi-Head Attention mechanism, which is a component that allows the TFT to focus on different segments of the input sequence. With multiple 'heads', the model can attend to various parts of the time series in parallel, granting it the ability to discern which historical factors are most relevant for predicting the future RCP. This attention to detail is crucial for tire health prediction, as it can reveal critical dependencies and relationships within the data—like the effects of temperature fluctuations or road conditions on tire wear—that might otherwise be overlooked.

Upon synthesizing the time series patterns and the critical insights from the GRN and the Masked Interpretable Multi-Head Attention mechanism, the tire health TFT channels this processed information into a Position-wise Feed-Forward Network. This final neural network layer is where all the preceding analytical components converge to produce the prediction of RCP.

For robust decision-making within our tire health TFT framework, it is essential to account for uncertainty in our predictions. To address this, we employ a Quantile loss function during the training phase of the model. This loss function allows us to estimate the expected value of the RCP at different quantiles, enabling the model to capture the variability and provide a more comprehensive range of potential outcomes. By training with this function, the model not only predicts a single point estimate but also provides a probabilistic understanding of tire health, thus significantly enhancing the reliability and robustness of the maintenance decisions based on these predictions[41]. The quantile loss is defined as:

$$L_q(y, \hat{y}) = q. \max(y - \hat{y}, 0) + (1 - q). \max(\hat{y} - y, 0) \qquad (7)$$

here, $L_q$ represents the quantile loss for a specified quantile $q$, $y$ is the actual value, $\hat{y}$ is the predicted value, $q$ specifies the quantile, and the equation calculates the loss by weighting underpredictions and overpredictions differently according to $q$.

The tire health TFT leverages Quantile Regression to quantify aleatoric uncertainty in tire wear predictions, producing forecasts across a range of outcomes from 0.01 to 0.99 quantiles. This approach not only enhances prediction accuracy for the RCP but also provides a statistical confidence interval, giving a comprehensive view of the predictive reliability and the inherent variability in the data.

Building on the quantification of aleatoric uncertainty through Quantile Regression, the tire health TFT also integrates Bayesian inference to measure epistemic uncertainty [42]. This is accomplished by implementing Monte Carlo dropout, a technique that simulates the effect of sampling from a probability distribution over the model's weights. By incorporating dropout layers during both the training and prediction phases, the model randomly omits subsets of connections, effectively generating a diverse set of outcomes with each forward pass [43]. The variance among these outcomes reflects epistemic uncertainty. The inclusion of Monte Carlo dropout enhances the model's ability to make informed decisions under uncertainty, which is particularly valuable in scenarios where data may be sparse or noisy. By running multiple forward passes and observing the variability in the results, we gain a more nuanced understanding of the prediction space, enabling more confident and reliable decision-making regarding tire maintenance. While making prediction the uncertainty from the quantile loss, and Bayesian inference are combined. The tire health TFT's training spans over 100 epochs, with a training and testing split of 80-20 %, ensuring a thorough learning process and robust convergence of the predictive model. The MAPE of the trained model is 29.5%. Following the detailed exploration of uncertainty quantification and decision-making mechanisms integrated into the tire health TFT (Temporal Fusion Transformer), a comparative study was conducted to evaluate its performance against other prominent sequence modeling techniques. The study examined the Mean Absolute Percentage Error (MAPE) across different models, demonstrating the effectiveness of the TFT. Below is a table showcasing the results:

*Table 1.  Comparative MAPE Performance of Sequence Modeling Techniques in Tire Health Prediction*

| Model | MAPE(%) |
|---|---|
| Long Short Term Memory (LSTM) | 36.23 |
| Recurrent Neural Network (RNN) | 37.85 |
| Gated Recurrent Unit (GRU) | 38.61 |
| Transformer Encoder | 35.40 |
| Temporal Fusion Transformer (TFT) | 29.50 |

Table 1 illustrates that the TFT model outperforms other models, including LSTM, RNN, GRU, and Transformer Encoder, by achieving the lowest MAPE value of 29.5%. This substantiates the TFT's ability in handling the complexities of predicting tire health, benefitting significantly from its advanced uncertainty quantification techniques and robust training regimen.

To ensure the tire health digital twin framework remains up-to-date and reflects the latest real-world conditions, we've integrated a discrepancy-based model update mechanism. This method evaluates the error between the model's forecasts and actual tire wear data, particularly valuable in scenarios where time series data is scarce or incomplete. By actively adjusting the tire health digital twin based on these discrepancies, we ensure that the digital twin continuously adapts, maintaining high accuracy in predicting the RCP despite the evolving nature of tire wear patterns. This model update method is discussed in the next subsection.

### 3.3 Model update with real-time data in tire health digital twin with discrepancy function

Moving forward we discuss the second important step in implementing the tire health digital twin which is model update.

Model update is one of the critical steps in any digital twin, as it keeps it's predictive capabilities similar to the real system [44]. In addressing the challenge of limited time series data for tire health digital twin updates we've implemented a discrepancy function-based model update approach. This method is centered on quantifying the deviation between the tire health TFT predictions and actual observations, an important step for environments where data scarcity could undermine model accuracy. The discrepancy calculation follows a straightforward equation,

$$D_{RCP}(t) = RCP_{real}(t) - RCP_{predicted}(t) \tag{8}$$

where $D_{RCP}(t)$ represents the discrepancy between the observed and predicted values at time $t$, $RCP_{real}(t)$ denotes the actual observed value at time $t$, $RCP_{predicted}(t)$ denotes the value predicted by the model at the same time point at time $t$.

Following this discrepancy calculation, a TFT model is fit on $D_{RCP}(t)$ with the same inputs used for the original $RCP_{predicted}(t)$ predictions. This ensures that the discrepancy model, henceforth referred to as Discrepancy $TFT$ ($DTFT$), is finely tuned to predict the difference between the actual and the initially predicted values from the tire health TFT model. The $DTFT's$ ability to forecast $D_{RCP}(t)$ allows for an adjustment of $RCP_{predicted}(t)$ towards $RCP_{real}(t)$. The input and output parameters of the DTFT model can be written as:

$$\widehat{D_{RCP}(t)} = DTFT(X(t), X'(t)) \tag{9}$$

Next, we will be calculating the corrected $RCP$, termed $RCP_{corrected}(t)$, by integrating the forecasted discrepancy from the $DTFT$ into the original $RCP_{predicted}(t)$. This is a hybrid modeling approach in which the tire health TFT predictions are improved by combining them with the DTFT predictions with respect to time [44]. This corrected $RCP$ is determined by applying the predicted discrepancy to the initial model predictions, thus:

$$RCP_{corrected}(t) = RCP_{predicted}(t) + \widehat{D_{RCP}}(t) \tag{10}$$

This equation ensures that the final predictions are adjusted to align with the actual observed values more closely, $RCP_{real}(t)$, thereby enhancing the accuracy and reliability of the digital twin's output. By leveraging the Discrepancy TFT model's predictions, our approach fine-tunes the digital twin's forecasts for tire health, enhancing predictive maintenance accuracy, especially in data-scarce situations, and improves as it assimilates more information.

### 3.4 Tire State Decision Algorithm for tire removal decision making.

As a next step in our Tire Health Digital Twin, we discuss the decision-making algorithm which is the tire state decision algorithm which will aid us in making the decision of when to change the tire. The objective of the tire health digital twin framework is to make tire replacement decision by identifying when a tire is likely to incur casing endurance damage. Casing endurance damage of a tire is defined as the degradation in the tire's structural integrity and performance beyond a certain threshold, which compromises its safety and effectiveness [45]. This determination is crucial for ensuring optimal tire life while preventing premature replacement and maintaining road safety [45].

To achieve this goal, we find a "removal from service" threshold for the RCP determined through the analysis of tire damage data and operational conditions which are collected by Michelin field test. This enables us to find the mileage beyond which the tire is likely to deplete its potential, thereby offering guidance for timely tire replacements to maximize safety and maintenance effectiveness.

Identifying the RCP threshold is a simple optimization step in which we select a variety of RCP values and determine the corresponding removal from service mileage using the Tire Health TFT. We then compare this predicted removal from service mileages to actual achieved mileages, which are collected from real-world tire usage data, using the Mean Absolute Percentage Error (MAPE) as our comparison metric. This approach allows us to accurately gauge the model's precision and adjust the removal from service threshold for reliable tire replacement predictions.

In Figure 4, we illustrate the relationship between the RCP threshold for tire removal from service, plotted on the x-axis, and the corresponding MAPE values, displayed on the y-axis. The objective of this analysis is to identify an RCP threshold where the MAPE is minimized, which would indicate the most accurate prediction for tire replacement. Through this minimization approach, we have identified 0.0909 as the optimal RCP threshold, corresponding to the lowest MAPE of 29.5%. Such a MAPE reflects a commendable precision level of our model's predictions, which is pivotal for making well-timed decisions about tire replacements. Achieving a MAPE of 29.5% underscores the model's efficacy in accurately forecasting tire health, thereby enabling precise, data-informed tire replacement strategies.

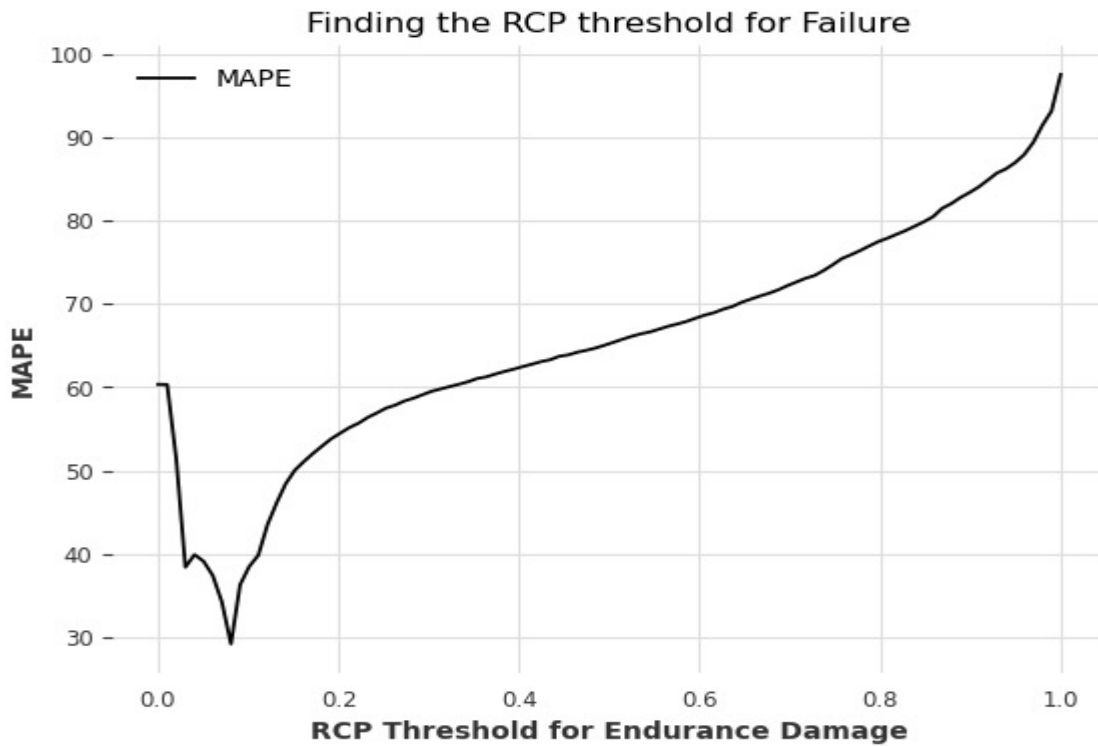

**FIGURE 4.** Optimization of RCP threshold for predicting tire removal mileage

Using the RCP threshold for removal from service, we employ a method named tire state decision algorithm that leverages recent tire health data, specifically over the last 20,000 kilometers to make a prediction of when to change in tires in mileage (kms). This method assumes that when the Tire RCP goes below 0.0909 the tire is likely to exhibit endurance damages.

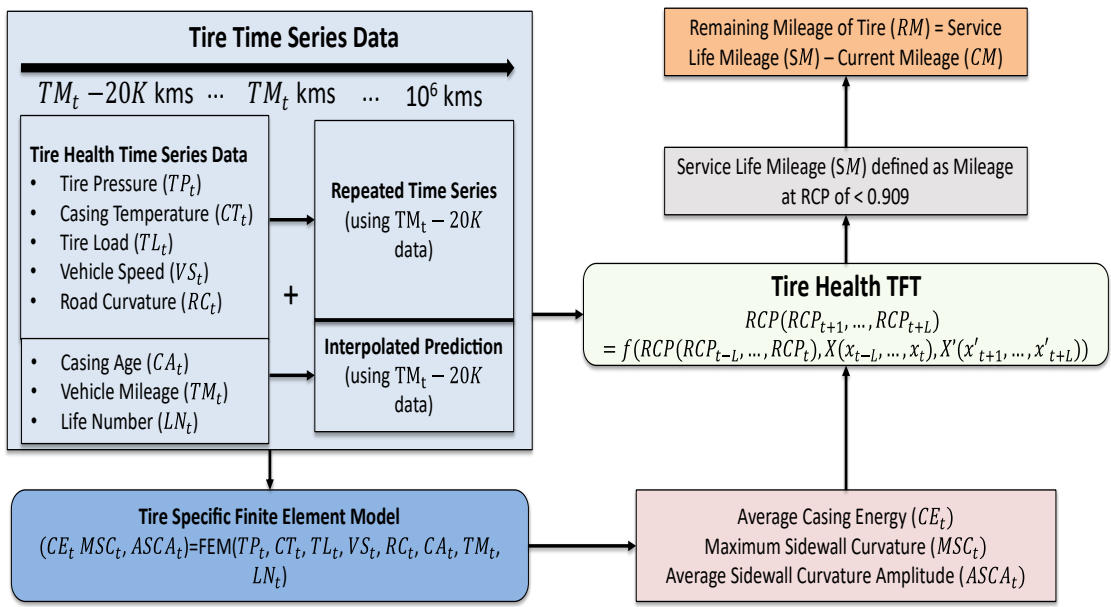

**FIGURE 5.** Tire state decision algorithm

To do so we repeat the time series of $TP_t$, $CT_t$, $TL_t$, $VS_t$, and $RC_t$, and we interpolate time series of $CA_t$, $TM_t$, and $LN_t$ as shown in Figure 5. The repetition and interpolation are referred from the previous 20,000 kms which can also be seen as $VM_t$-20,000 kms, and then repeated and interpolated until the tire mileage reaches 1 million kms. The seven input variables are processed through a tire-specific Finite Element Model ($FEM$) to yield three critical outputs: Average Casing Energy ($CE_t$), reflecting the cumulative stress and wear on the tire; Maximum Sidewall Curvature ($MSC_t$), indicating peak sidewall bending under load; and Average Sidewall Curvature Amplitude ($ASCA_t$), measuring cyclic fatigue stress. These metrics collectively provide a comprehensive view of the tire's condition and endurance life. Now, we input this data to the tire health TFT, and then find the removal mileage of the Tire as shown in Figure 5. By systematically applying this method throughout the tire's service life, the tire state decision algorithm provides guidance on the optimal mileage to schedule tire replacement, as illustrated in Figure 5. Moving forward, we will be discussing the results achieved from the Tire health Digital Twin.

### 4. RESULTS AND DISCUSSION

In this section, we investigate the RCP predictions by the tire health digital twin model, focusing on a comparison between two specific tires and their respective RCP projections within two separate tires location and vehicle fleets. Additionally, we see the results from the model update in which we update the tire health digital twin on data which has high temperature by using the discrepancy update method. Additionally, we discuss the results from the Tire State decision algorithm in which we observe how the mileage before failure changes at different vehicle states.

### 4.1 Validation of the tire health TFT

In this section, the Tire Health TFT model is validated across a range of scenarios to evaluate its predictive performance. This approach ensures that the model's behavior remains consistent and reliable under various conditions, a crucial step for validating its effectiveness in real-world applications.

First, we compare tires at different Tire axle location on the truck which are the steer location and drive location. As shown in Figure 6 it is evident that tires positioned at the steer axle are more likely subject to endurance damage earlier than those located at the drive axle. The prediction from tire health DT affirms that steer tires, even when maintained at the same pressure as drive tires, tend to exhibit endurance damages sooner. The reason for this accelerated wear pattern lies in the specific demands placed on steer tires. Unlike the drive tires, which carry a load that is distributed over multiple axles and sometime in a dual wheel configuration, steer axle is only made of 2 tires, and each is usually subject to higher load and a variety of other stresses. When a vehicle turns or maneuvers, the steer tires must handle dynamic forces such as lateral pressure, which can significantly affect their structural integrity over time. This distinction in stress responses between tire types is crucial for understanding tire longevity. Recognizing this unique endurance damage is essential in a predictive maintenance framework, as it allows for more accurate forecasting of tire life and timely maintenance actions. These observations demonstrate the diverse damage occurrence patterns of tires based on their position on a vehicle and demonstrate the accuracy of the Tire Health DT in capturing these patterns to predict tire health across different axle locations.

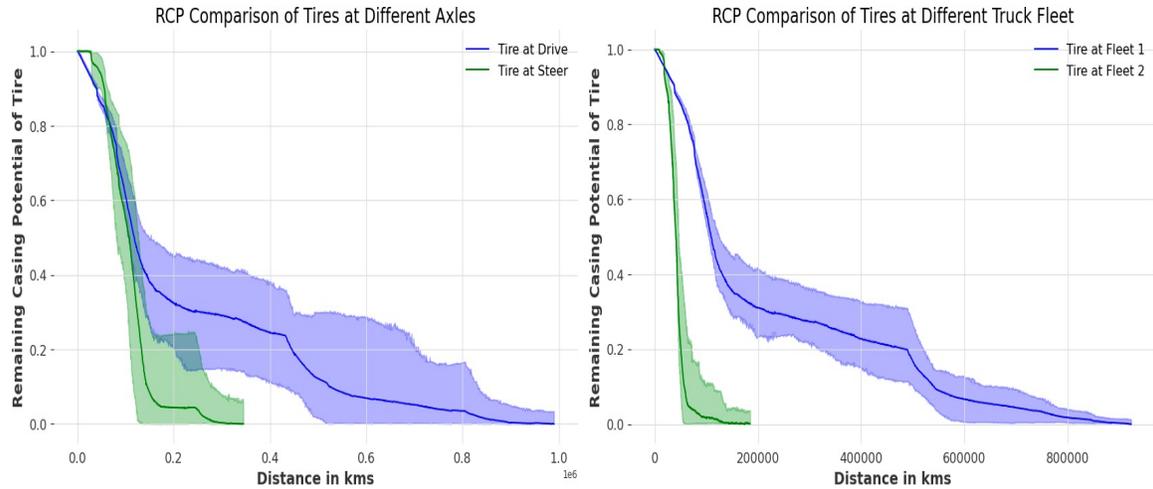

**FIGURE 6.** Comparative analysis of RCP for tires on different axes and fleet trucks: the left plot illustrates the RCP trajectories for two tires situated on different axles, showcasing the variances in wear patterns between steer and drive positions. The right plot contrasts the RCP of tires from two separate truck fleets, highlighting the impact of fleet-specific operational conditions on tire longevity. The shaded regions represent 95 % confidence intervals, providing a visual measure of prediction certainty over the distance covered.

In our second validation analysis, we extend the scope of our validation of tire health TFT by including time series data from tires of two separate truck fleets. Truck fleet 1 tires are characterized by lower-than-recommended pressure levels and higher temperatures, while truck fleet 2 tires maintain standard pressure and temperature conditions. As shown in Figure 6 tires from Fleet 1 demonstrate a quicker progression to endurance damage compared to those in Fleet 2. This finding correlates with the well-established principle in tire mechanics that tires operating under conditions of low pressure and high temperature are more susceptible to rapid wear and potential endurance damage. The accelerated degradation of tire health is a consequence of the tire's compromised structural integrity when subjected to such stresses. These conditions often lead to increased friction and deformation of the tire material, which, over time, can significantly shorten the tire's serviceable life.

The comparative results validate the accuracy of our Tire Health DT, demonstrating its ability to predict endurance damage based on different tire positions on a truck axle and various truck fleet operating conditions. This validation highlights the model's capabilities as a dependable instrument for predicting tire health across a wide array of real-world situations. Such validation is essential, confirming the model's applicability for maintenance planners seeking to optimize tire life and fleet performance.

**4.2 Updating the tire health DT to high-temperature data through discrepancy function approach**

In this section, we explore the outcomes when the Tire Health TFT is applied to a new dataset beyond the range of the training dataset, alongside the effectiveness of our discrepancy-based model updating technique in enhancing the digital twin framework's accuracy. These results provide insight into the digital twin's adaptability and the robustness of continuous update methods for predictive analytics in tire health monitoring.

For this section, we use time series dataset of tire health, specifically focusing on performance at elevated temperatures, recorded at a notable 100 ℃. This data is collected from the controlled drum test environment provided by Michelin. As depicted in Figure 7, we noted a discrepancy in prediction data from the tire health TFT when tires were exposed to the extreme temperature of 100 °C. The actual tire endurance damage occurred at 110,000 kms, which is a contrast to the 321,000 kms forecasted by our tire health TFT model which is the surrogate model of digital twin. This discrepancy underscores the significant influence that elevated temperatures have on tire life and the resultant complications they introduce to the precision of tire health analysis. The reason for the discrepancy observed is due to the offline training of the tire health TFT model on the historical tire health data, which primarily utilized datasets encompassing a temperate range from 0 °C to 50 °C. This shows that the model's predictive robustness is challenged when faced with the extremes of high-temperature conditions that exceed its initial training dataset, highlighting an area for further update in the digital twin framework to enhance its applicability across a broader range of operational environments.

To improve the model's adaptability to dataset it is not trained on, a discrepancy-based method is utilized, adjusting the predictions by incorporating high-temperature drum test data where tires reached their potential prematurely. As shown in figure 7, the prediction from the tire health TFT is added with the prediction from the DTFT, to give the corrected prediction. The corrected prediction of the RCP from the fused model is much closer to the real prediction, where according to the fused model the endurance damage occurs in the tire at 121,000 kms which is closer to our observation of 111,000 kms. Improving the tire health digital twin's accuracy for tire health monitoring is of paramount importance, especially when dealing with operational scenarios beyond its initial training process. By integrating a discrepancy-based approach, the tire health digital twin is fine-tuned to factor in data from high-temperature scenarios that cause an accelerated endurance damage occurrence. This method of model updating is important as it ensures the tire health monitoring system remains robust and reliable across varying and unforeseen conditions, thus safeguarding against premature tire degradation and ensuring vehicular safety. In the next section, we will be discussing the third important pillar of a digital twin framework which is tire state decision method.

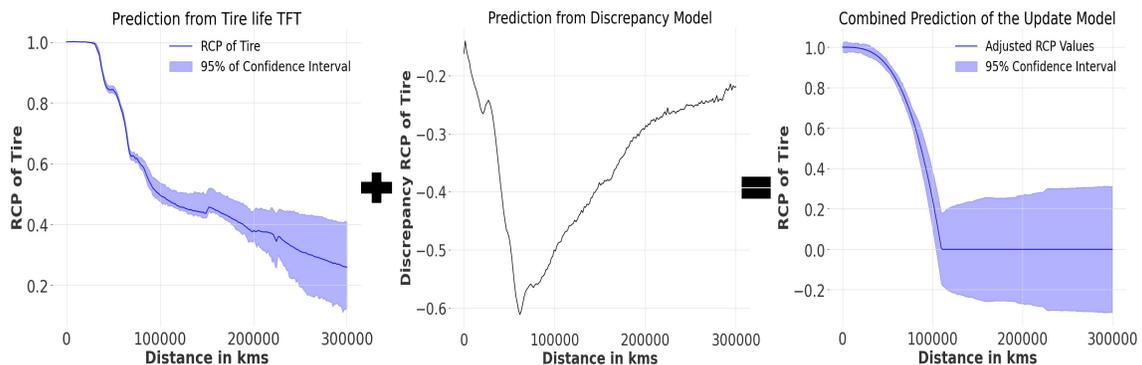

**FIGURE 7.** The first graph showcases the RCP forecasted by the tire health TFT model, with a 95% confidence interval. The middle graph represents the predictions from the discrepancy model (DTFT), which predicts the deviation between observed and predicted RCP. The final graph on the right combines these two models to provide a unified prediction of tire life, encapsulating both the adjusted RCP values and the associated confidence intervals.

### 4.3 Use case for tire state decision method

In this section, we study the practical applications of our tire health digital twin framework, aimed at predicting the RCP and remaining mileage of the tire at different distance intervals in service in life. We demonstrate the tire state decision method as a feature for the users to check the remaining mileage before the tire has an endurance damage at different time steps. For this paper we predict the remaining mileage before the tire has an endurance damage at 40,000, 82,000, 124,000, 166,000, 208,000, and 250,000 kilometers as shown in Figure 8. The figure represents these predictions, showing the RCP against distance traveled, underscored by a 95% confidence interval that captures the predictive uncertainty. The future tire health data is created according to method discussed in section 3.4.

Initially, the predictions by our tire health digital twin framework showed a certain degree of inaccuracy, as highlighted by the broader confidence intervals in the early stages of the tire's life depicted in the Figure 8. However, as more mileage data became available, particularly at higher distances, the model's predictions converged more closely with the actual removal mileage. The decision-making algorithm for tire removal is a crucial component in our tire health digital twin framework, offering significant enhancements in accuracy over the tire's lifespan. This algorithm becomes particularly valuable as it adjusts and refines its predictions with the accumulation of real-world mileage data, effectively narrowing the uncertainty margins and aligning its outputs with the actual tire removal milestones. This refined accuracy is essential for operators, allowing for well-informed maintenance decisions that optimize tire usage and ensure safety.

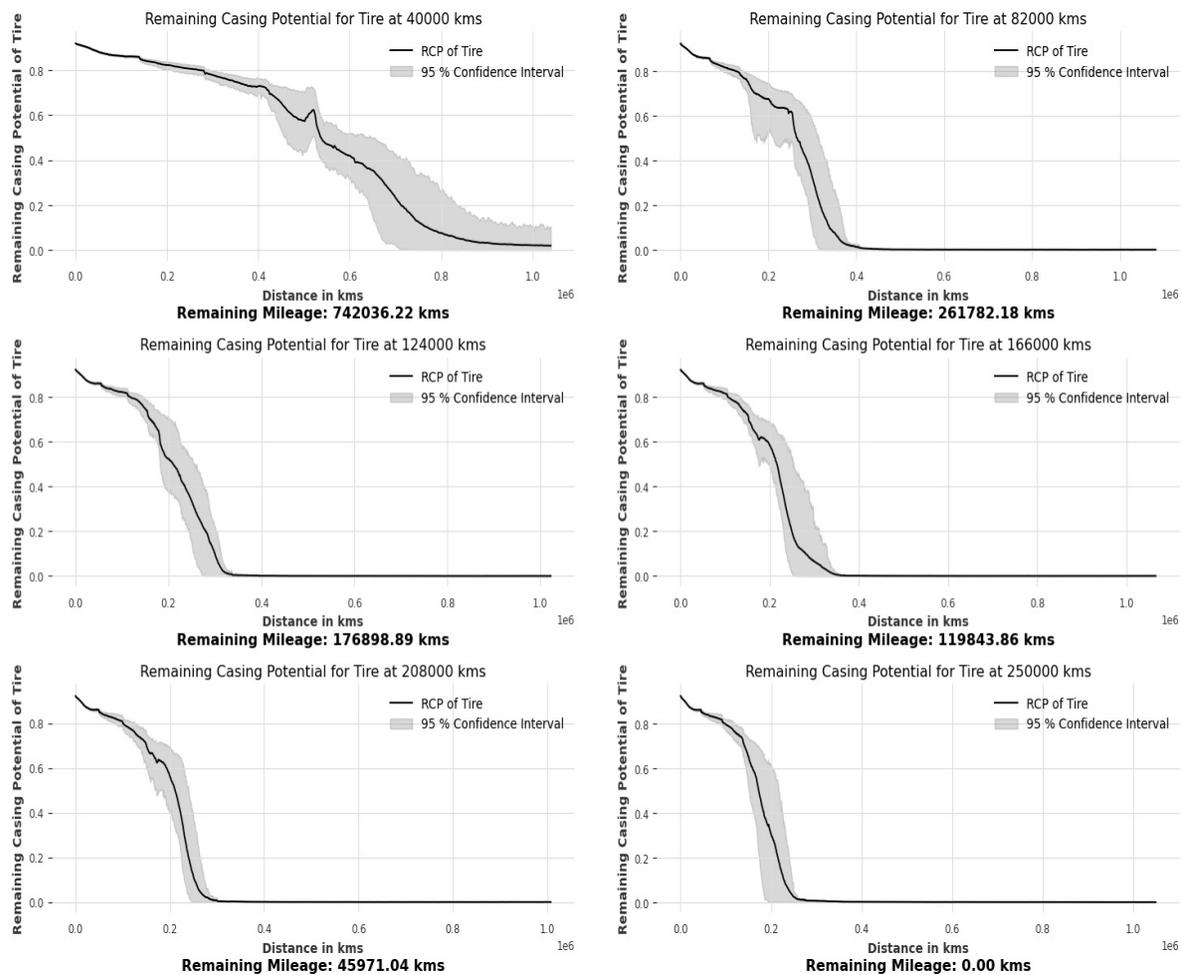

**FIGURE 8.** Series of plots illustrating the RCP for a tire at various mileage intervals. Each plot shows the predicted RCP the different tire life instances, with a shaded area representing the 95% confidence interval around the prediction.

## 5. CONCLUSION

In this paper, we have addressed challenges in leveraging digital twin technology for predictive maintenance, specifically focusing on tire health prediction. By detailing the implementation of a Tire health digital twin, we demonstrated how we deal with tire health data with damage which is difficult to measure in real-time. As our first step in building a tire health digital twin we demonstrate the effective offline training of a Tire Health TFT model by combining operating, usage, and state variables to predict the health of the tire, emphasizing the importance of accurately quantifying both epistemic and aleatoric uncertainty within our predictions. Furthermore, as a second step to build our tire health digital twin, we demonstrate the model's update mechanism utilizing the discrepancy update method, enabling digital twin refinement even with limited tire damage data, thereby enhancing the accuracy and effectiveness of our tire health digital twin. As a third step in developing our digital twin, we introduce a tire state decision method, which assists in making decision of the optimal time to replace a tire for maintenance purposes. Our methodology for updating the tire health digital twin based on discrepancy method, along with our strategies for minimizing uncertainty in decision-making, illustrates the significant advantages digital twins offer for predictive maintenance practices. Through this digital twin framework, we contribute to the field of predictive maintenance by providing a solution for extending tire life and enhancing vehicle safety through data-driven insights.

**ACKNOWLEDGMENT**

Grant support from the Remade institute research program

(DE-EE0007897) is greatly acknowledged. The authors would also like to acknowledge support from the NSF Engineering Research Center for Hybrid Autonomous Manufacturing Moving from Evolution to Revolution (ERC-HAMMER) under Award Number EEC-2133630.

**APPENDIX**

Definition of the variables used in the tire health digital twin are as follows:

*Tire Mileage ($TM_t$):* Total distance covered by the vehicle up to time $t$, affecting the tire's wear and health.

*Casing Age ($CA_t$):* Time since the tire's manufacture, indicating potential degradation due to aging.

*Life Number ($LN_t$):* Indicates the number of retreads a tire has undergone, with 1 representing new casings.

*Tire Load* ($LT_t$)*:* The load carried by the tire, varying based on wheel position but consistent over long journeys.

*Vehicle Speed* ($VS_t$)*:* The speed at which the vehicle operates.

*Casing Temperature* ($CT_t$)*:* Temperature inside the tire, varying with wheel position and external conditions.

*Tire Pressure* ($TP_t$)*:* Pressure within the tire, changing with location on the vehicle and internal temperature fluctuations.

*Average Casing Energy* ($CE_t$)*:* The cumulative elastic energy absorbed by the tire's casing throughout its life. This energy is calculated via finite element models under known conditions of pressure, load, and speed, then aggregated across each tire rotation. This metric reflects the overall stress and strain endured by the tire, contributing to its wear and performance characteristics.

*Maximum Sidewall Curvature* ($MSC_t$)*:* The peak curvature experienced by the tire's sidewall in a single rotation, influenced by the tire's load and pressure settings. Higher loads and lower pressures increase sidewall curvature, potentially reducing the fatigue life of the tire's cables.

*Average Sidewall Curvature Amplitude* ($ASCA_t$)*:* The mean difference between the maximum and minimum sidewall curvatures during a rotation, related to the bending stress amplitude. This metric is indicative of cyclic fatigue experienced by the tire's cables.

*Road curvature of the Vehicle* ($RC_t$): Calculated from the vehicle acceleration and speed, indicating the tire's average exposure to turning forces. Lower values typically occur in urban environments due to tighter turns, whereas higher values are more common on highways.

**REFERENCES**

[1]  Weissman, S., Sackman, J., Gillen, D., and Monismith, C., "Extending the Lifespan of Tires: Final Report."

[2]  "The Role of Data Fusion in Predictive Maintenance Using Digital Twin | AIP Conference Proceedings | AIP Publishing" [Online]. Available: https://pubs.aip.org/aip/acp/article/1949/1/020023/819472/The-role-of-data-fusion-in-predictive-maintenance. [Accessed: 04-Mar-2024].

[3]  Fuller, A., Fan, Z., Day, C., and Barlow, C., 2020, "Digital Twin: Enabling Technologies, Challenges and Open Research," IEEE Access, **8**, pp. 108952–108971.

[4]  Hummen, T., Hellweg, S., and Roshandel, R., 2023, "Optimizing Lifespan of Circular Products: A Generic Dynamic Programming Approach for Energy-Using Products," Energies, **16**(18), p. 6711.

[5]  Nouri Qarahasanlou, A., Ataei, M., Khalokakaie, R., Ghodrati, B., and Jafarei, R., 2016, "Tire Demand Planning Based on Reliability and Operating Environment," Int. J. Min. Geo-Eng., **50**(2), pp. 239–248.

[6]  Jha, D., Karkaria, V. N., Karandikar, P. B., and Desai, R. S., 2022, "Statistical Modeling of Hybrid Supercapacitor," J. Energy Storage, **46**, p. 103869.

[7]  Symeonides, D., Loizia, P., and Zorpas, A. A., 2019, "Tire Waste Management System in Cyprus in the Framework of Circular Economy Strategy," Environ. Sci. Pollut. Res., **26**(35), pp. 35445–35460.

[8]  Gehrke, I., Schläfle, S., Bertling, R., Öz, M., and Gregory, K., 2023, "Review: Mitigation Measures to Reduce Tire and Road Wear Particles," Sci. Total Environ., **904**, p. 166537.

[9]  Karkaria, V., Chen, J., Siuta, C., Lim, D., Radelescu, R., and Chen, W., 2024, "Tire Life Assessment for Increasing Re-Manufacturing of Commercial Vehicle Tires," *Technology Innovation for the Circular Economy*, John Wiley & Sons, Ltd, pp. 599–612.

[10] Karkaria, V., Chen, J., Siuta, C., Lim, D., Radulescu, R., and Chen, W., 2023, "A Machine Learning–Based Tire Life Prediction Framework for Increasing Life of Commercial Vehicle Tires," J. Mech. Des., **146**(020902).

[11] Rahimdel, M. J., 2023, "Residual Lifetime Estimation for the Mining Truck Tires," Proc. Inst. Mech. Eng. Part J. Automob. Eng., **237**(13), pp. 3232–3244.

[12] *Read "Tires and Passenger Vehicle Fuel Economy: Informing Consumers, Improving Performance -- Special Report 286" at NAP.Edu*.

[13] Kong, T., Hu, T., Zhou, T., and Ye, Y., 2021, "Data Construction Method for the Applications of Workshop Digital Twin System," J. Manuf. Syst., **58**, pp. 323–328.

[14] Hussain, S., Mahmud, U., and Yang, S., 2021, "Car E-Talk: An IoT-Enabled Cloud-Assisted Smart Fleet Maintenance System," IEEE Internet Things J., **8**(12), pp. 9484–9494.

[15] Kapteyn, M. G., Knezevic, D. J., and Willcox, K., "Toward Predictive Digital Twins via Component-Based Reduced-Order Models and Interpretable Machine Learning," *AIAA Scitech 2020 Forum*, American Institute of Aeronautics and Astronautics.

[16] Bosso, N., Magelli, M., Trinchero, R., and Zampieri, N., 2024, "Application of Machine Learning Techniques to Build Digital Twins for Long Train Dynamics Simulations," Veh. Syst. Dyn., **62**(1), pp. 21–40.

[17] David, I., Galasso, J., and Syriani, E., 2021, "Inference of Simulation Models in Digital Twins by Reinforcement Learning," *2021 ACM/IEEE International Conference on Model Driven Engineering Languages and Systems Companion (MODELS-C)*, pp. 221–224.

[18] "Scientific Machine Learning Enables Multiphysics Digital Twins of Large-Scale Electronic Chips | IEEE Journals & Magazine | IEEE Xplore" [Online]. Available: https://ieeexplore.ieee.org/abstract/document/9924250. [Accessed: 11-May-2024].

[19] Tripura, T., Desai, A. S., Adhikari, S., and Chakraborty, S., 2023, "Probabilistic Machine Learning Based Predictive and Interpretable Digital Twin for Dynamical Systems," Comput. Struct., **281**, p. 107008.

[20] van Beek, A., Nevile Karkaria, V., and Chen, W., 2023, "Digital Twins for the Designs of Systems: A Perspective," Struct. Multidiscip. Optim., **66**(3), p. 49.

[21] Löcklin, A., Müller, M., Jung, T., Jazdi, N., White, D., and Weyrich, M., 2020, "Digital Twin for Verification and Validation of Industrial Automation Systems – a Survey," *2020 25th IEEE International Conference on Emerging Technologies and Factory Automation (ETFA)*, pp. 851–858.

[22] Liu, M., Fang, S., Dong, H., and Xu, C., 2021, "Review of Digital Twin about Concepts, Technologies, and Industrial Applications," J. Manuf. Syst., **58**, pp. 346–361.

[23] Mao, H., Liu, Z., Qiu, C., Huang, Y., and Tan, J., 2023, "Prescriptive Maintenance for Complex Products with Digital Twin Considering Production Planning and Resource Constraints," Meas. Sci. Technol., **34**(12), p. 125903.

[24] Li, Z., Wallace, E., Shen, S., Lin, K., Keutzer, K., Klein, D., and Gonzalez, J., 2020, "Train Big, Then Compress: Rethinking Model Size for Efficient Training and Inference of Transformers," *Proceedings of the 37th International Conference on Machine Learning*, PMLR, pp. 5958–5968.

[25] Hu, J., and Chen, P., 2020, "Predictive Maintenance of Systems Subject to Hard Failure Based on Proportional Hazards Model," Reliab. Eng. Syst. Saf., **196**, p. 106707.

[26] Zhu, J., Han, K., and Wang, S., 2021, "Automobile Tire Life Prediction Based on Image Processing and Machine Learning Technology," Adv. Mech. Eng., **13**(3), p. 16878140211002727.

[27] Lim, B., Arık, S. Ö., Loeff, N., and Pfister, T., 2021, "Temporal Fusion Transformers for Interpretable Multi-Horizon Time Series Forecasting," Int. J. Forecast., **37**(4), pp. 1748–1764.

[28] "A Machine Learning–Based Tire Life Prediction Framework for Increasing Life of Commercial Vehicle Tires | J. Mech. Des. | ASME Digital Collection" [Online]. Available: https://asmedigitalcollection.asme.org/mechanicaldesign/article/146/2/020902/1169300. [Accessed: 12-Feb-2024].

[29] Moges, T., Yang, Z., Jones, K., Feng, S., Witherell, P., and Lu, Y., 2021, "Hybrid Modeling Approach for Melt-Pool Prediction in Laser Powder Bed Fusion Additive Manufacturing," J. Comput. Inf. Sci. Eng., **21**(050902).

[30] Arendt, P. D., Apley, D. W., and Chen, W., 2012, "Quantification of Model Uncertainty: Calibration, Model Discrepancy, and Identifiability," J. Mech. Des., **134**(100908).

[31] Hickey, J., and Langley, R., 2022, "Alternative Metrics for Design Decisions Based on Separating Aleatory and Epistemic Probabilistic Uncertainties," Mech. Syst. Signal Process., **181**, p. 109532.

[32] Gal, Y., and Ghahramani, Z., 2016, "Dropout as a Bayesian Approximation: Representing Model Uncertainty in Deep Learning," *Proceedings of The 33rd International Conference on Machine Learning*, PMLR, pp. 1050–1059.

[33] Zhang, W., Quan, H., and Srinivasan, D., 2019, "An Improved Quantile Regression Neural Network for Probabilistic Load Forecasting," IEEE Trans. Smart Grid, **10**(4), pp. 4425–4434.

[34] "Known, Unknown, and Unknowable Uncertainties | Theory and Decision" [Online]. Available: https://link.springer.com/article/10.1023/A:1015544715608. [Accessed: 06-Mar-2024].

[35] Zhou, L., Pan, S., Wang, J., and Vasilakos, A. V., 2017, "Machine Learning on Big Data: Opportunities and Challenges," Neurocomputing, **237**, pp. 350–361.

[36] Chen, D., Hu, Q., and Yang, Y., 2011, "Parameterized Attribute Reduction with Gaussian Kernel Based Fuzzy Rough Sets," Inf. Sci., **181**(23), pp. 5169–5179.

[37] "Dimension Reduction With Extreme Learning Machine | IEEE Journals & Magazine | IEEE Xplore" [Online]. Available: https://ieeexplore.ieee.org/abstract/document/7471467. [Accessed: 04-Mar-2024].

[38] Xu, D., Wang, P., Jiang, Y., Fan, Z., and Wang, Z., 2022, "Signal Processing for Implicit Neural Representations," Adv. Neural Inf. Process. Syst., **35**, pp. 13404–13418.

[39] Zhang, Y., Wang, Y., Jiang, Z., Zheng, L., Chen, J., and Lu, J., 2023, "Domain Adaptation via Transferable Swin Transformer for Tire Defect Detection," Eng. Appl. Artif. Intell., **122**, p. 106109.

[40] Zhang, H., Zou, Y., Yang, X., and Yang, H., 2022, "A Temporal Fusion Transformer for Short-Term Freeway Traffic Speed Multistep Prediction," Neurocomputing, **500**, pp. 329–340.

[41] Xu, Q., Deng, K., Jiang, C., Sun, F., and Huang, X., 2017, "Composite Quantile Regression Neural Network with Applications," Expert Syst. Appl., **76**, pp. 129–139.

[42] Sadr, M. A. M., Gante, J., Champagne, B., Falcao, G., and Sousa, L., 2022, "Uncertainty Estimation via Monte Carlo Dropout in CNN-Based mmWave MIMO Localization," IEEE Signal Process. Lett., **29**, pp. 269–273.

[43] Karkaria, V., Goeckner, A., Zha, R., Chen, J., Zhang, J., Zhu, Q., Cao, J., Gao, R. X., and Chen, W., 2024, "Towards a Digital Twin Framework in Additive Manufacturing: Machine Learning and Bayesian Optimization for Time Series Process Optimization."

[44] Zhang, H., Qi, Q., Ji, W., and Tao, F., 2023, "An Update Method for Digital Twin Multi-Dimension Models," Robot. Comput.-Integr. Manuf., **80**, p. 102481.

[45] Nyaaba, W., Frimpong, S., and Anani, A., 2019, "Fatigue Damage Investigation of Ultra-Large Tire Components," Int. J. Fatigue, **119**, pp. 247–260.

**Figure Captions List**

Fig. 1    Tire health digital twin framework demonstrating the flow of information, and important components of digital twin like offline training, model update and decision making. The operating, usage and state parameters $(T_t, U_t, M_t)$ collectively form the input variable $(X)$ for the tire health TFT model.

Fig. 2    Flowchart depicting the causal relationships and information flow among input variables leading to the tire health TFT output. The diagram illustrates the hierarchically structured data inputs, from tire mileage to individual tire factors, culminating in the prediction of RCP.

Fig. 3    Architecture of the tire health TFT model. It illustrates the multilayered flow of information within the model, starting from variable selection at the lowest level, ascending through the LSTM encoder-decoder structure for temporal analysis, and culminating in the dense and GRN layers that synthesize the input to predict the RCP.

Fig. 4    Optimization of RCP threshold for predicting tire removal mileage

Fig. 5    Tire state decision algorithm

Fig. 6    Comparative analysis of RCP for tires on different axes and fleet trucks: the left plot illustrates the RCP trajectories for two tires situated on different axles, showcasing the variances in wear patterns between steer and drive positions. The right plot contrasts the RCP of tires from two

separate truck fleets, highlighting the impact of fleet-specific operational conditions on tire longevity. The shaded regions represent 95 % confidence intervals, providing a visual measure of prediction certainty over the distance covered.

Fig. 7     The first graph showcases the RCP forecasted by the tire health TFT model, with a 95% confidence interval. The middle graph represents the predictions from the discrepancy model (DTFT), which predicts the deviation between observed and predicted RCP. The final graph on the right combines these two models to provide a unified prediction of tire life, encapsulating both the adjusted RCP values and the associated confidence intervals.

Fig. 8     Series of plots illustrating the RCP for a tire at various mileage intervals. Each plot shows the predicted RCP the different tire life instances, with a shaded area representing the 95% confidence interval around the prediction.

**Table Caption List**

Table 1        Comparative MAPE Performance of Sequence Modeling Techniques in Tire Health Prediction